\documentclass[10pt,twocolumn,letterpaper]{article}

\usepackage[utf8]{inputenc}
\usepackage{3dv}
\usepackage{times}
\usepackage{epsfig}
\usepackage{graphicx,subcaption}
\usepackage{amsmath}
\usepackage{amssymb}
\usepackage{makecell}
\usepackage{arydshln}
\usepackage{gensymb}
\usepackage{pdfpages}
\usepackage{booktabs}
\usepackage{stmaryrd}

\newcommand{\ra}[1]{\renewcommand{\arraystretch}{#1}}

\usepackage{multirow}
\usepackage{array}
\newcolumntype{?}{!{\vrule width 1pt}}
\usepackage[square,numbers,sort]{natbib}
\usepackage{xcolor}

\newcommand{\commented}[1]{}
\newcommand\boldblue[1]{\textcolor{blue}{\textbf{#1}}}
\newcommand\boldred[1]{\textcolor{red}{\textbf{#1}}}
\newcommand{\code}{\texttt}

\newcommand{\vincent}[1]{#1}
\newcommand{\vincentrmk}[1]{}
\newcommand{\hugo}[1]{#1}
\newcommand{\hugormk}[1]{}

\newcommand{\gbrmk}[1]{}

\usepackage[pagebackref=true,breaklinks=true,urlcolor=blue,letterpaper=true,colorlinks,bookmarks=false]{hyperref}
\threedvfinalcopy 
\def\threedvPaperID{88} 

\ifthreedvfinal\fi
\begin{document}

\title{Sparse-to-Dense Hypercolumn Matching for Long-Term Visual Localization}
\author{Hugo Germain\footnotemark[1]~~\textsuperscript{1} \hspace{1em}
  Guillaume Bourmaud\footnotemark[1]~~\textsuperscript{2} \hspace{1em}
  Vincent Lepetit\footnotemark[1]~~\textsuperscript{1} \\
  \textsuperscript{1}Laboratoire Bordelais de Recherche en Informatique, Université de Bordeaux, France\\
  \textsuperscript{2}Laboratoire IMS, Université de Bordeaux, France\\
  }
\maketitle
\thispagestyle{plain}
\pagestyle{plain}
\setcounter{footnote}{1}\footnotetext{E-mail: {\tt\small \{firstname.lastname\}@u-bordeaux.fr}}

\begin{abstract}

We propose a  novel approach to feature point matching,  suitable for robust and
accurate  outdoor visual  localization in  long-term scenarios.   Given a  query
image, we  first match  it against  a database  of registered  reference images,
using recent retrieval techniques. This gives  us a first estimate of the camera
pose.  To  refine this estimate,  like previous  approaches, we match  2D points
across the query  image and the retrieved reference image.   This step, however,
is  prone to  fail as  it is  still very  difficult to  detect and  match sparse
feature points across images captured  in potentially very different conditions.
Our key contribution  is to show that  we need to extract  sparse feature points
only in the  retrieved reference image: We then search  for the corresponding 2D
locations  in  the query  image  exhaustively.   This  search can  be  performed
efficiently using  convolutional operations,  and robustly by  using hypercolumn
descriptors, \emph{i.e.}   image features computed  for retrieval.  We  refer to
this method as  'Sparse-to-Dense Hypercolumn Matching'.  Because we  know the 3D
locations of  the sparse  feature points  in the reference  images thanks  to an
offline reconstruction  stage, it  is then possible  to accurately  estimate the
camera pose from these matches.  Our experiments show that this method allows us
to outperform the state-of-the-art on several challenging outdoor datasets.

\end{abstract}

\section{Introduction}

Visual localization  is a key  component to  many robotic systems,  ranging from
autonomous   navigation~\cite{McManus2014ShadyDR}   to    augmented   or   mixed
reality~\cite{Middelberg2014Scalable6L}.  Yet,  accurately predicting the  6 DoF
camera pose of a visual query with  respect to a reference frame can become very
challenging  in  long-term  scenarios:  Despite recent  progress,  many  outdoor
location methods are still prone to fail especially at high precision thresholds
and under day-to-night changes~\cite{6DOFBenchmark} as images can undergo a wide
variety of visual changes between different time of day and across seasons.

\begin{figure}
  \begin{center}
    \begin{subfigure}[t]{0.9\columnwidth}
      \centering\includegraphics[width=0.49\columnwidth]{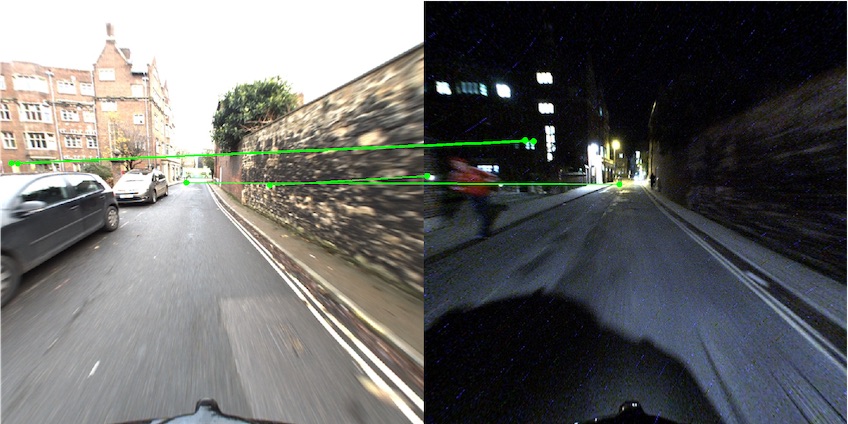}
      \centering\includegraphics[width=0.49\columnwidth]{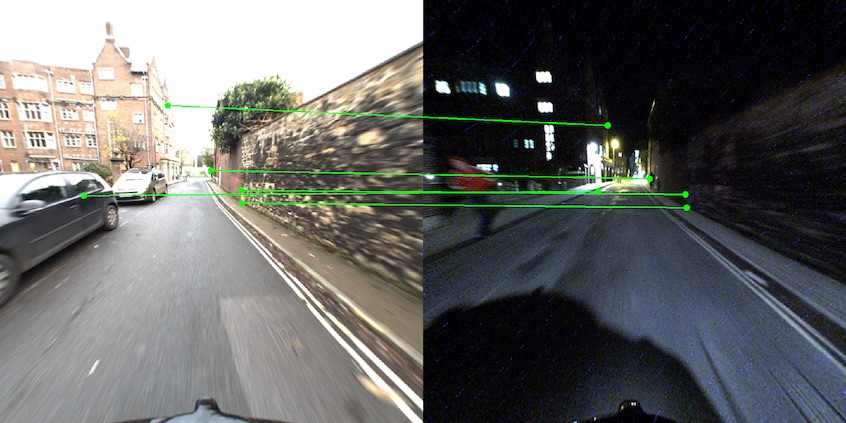}
      \caption{Standard sparse-to-sparse matching using SuperPoint detections and two different descriptors : (Left) SuperPoint descriptors [4 inliers], (Right) HyperColumn descriptors [5 inliers]}
    \end{subfigure}
    ~
    \begin{subfigure}[t]{0.9\columnwidth}
      \centering\includegraphics[width=\columnwidth]{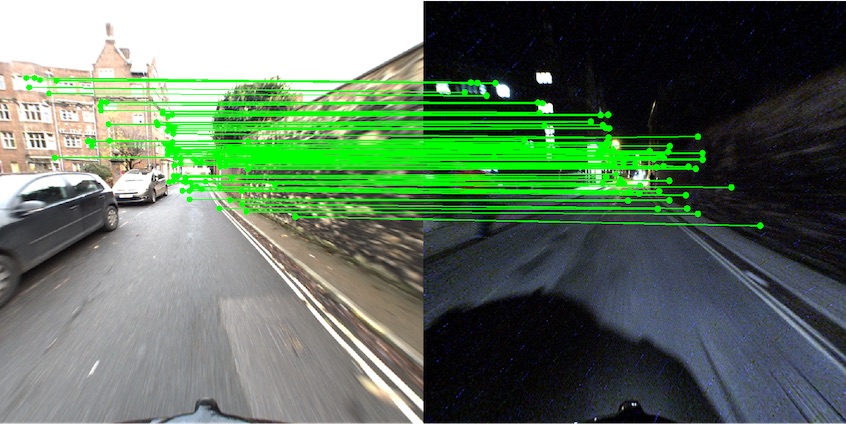}
      \caption{Our sparse-to-dense matching approach using Superpoint detections in  the left image only and HyperColumn
        descriptors [87 inliers]}
    \end{subfigure}
  \end{center}
  \vspace{-0.5cm}
  \caption{Top images: Despite recent progress, matching sparse feature points extracted from two images captured under very different conditions remains extremely challenging.  Bottom image: Our key contribution is to show that it is much more robust to extract sparse feature points in only one image, and to search for their correspondents exhaustively in the other image.  This exhaustive search can be performed very efficiently using convolutional operations.  Using the 3D locations of the sparse feature points, we can then compute the camera pose. We show the number of inlier matches found by PnP+RANSAC.}\label{fig:ablation}
\end{figure}


\begin{figure*}[t]
  \begin{center}
    \includegraphics[width=0.95\linewidth]{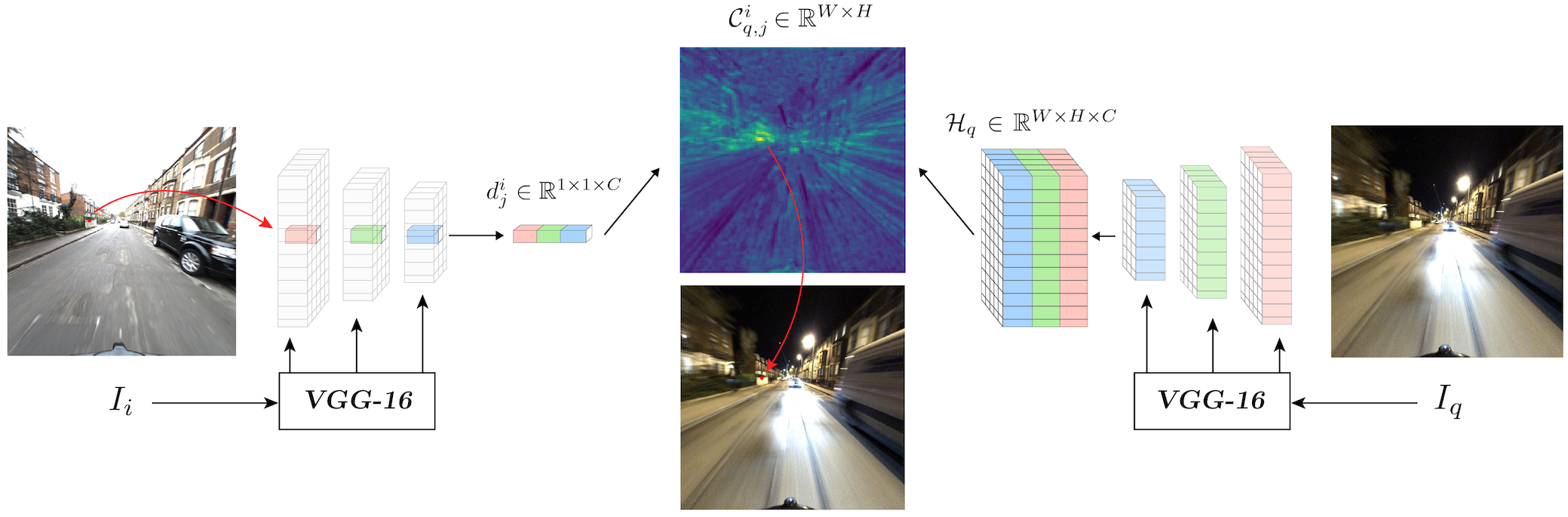}
  \end{center}
    \vspace{-2mm}
  \caption{\textbf{Sparse-to-dense feature matching using hypercolumns.} For
    each detection $m_j^i$in the reference image $I_i$ retrieved for query image
    $I_q$, we extract a hypercolumn descriptor $d_j^i$, which we cross-correlate
    exhaustively against the dense hypercolumn $\mathcal{H}_q$. We then define
    the correspondent location of $m_j^i$ in $I_q$ as the image location of the
    maximum value in the resulting correlation map $\mathcal{C}_{q,j}^i = d_j^i
    * \mathcal{H}_q$.}\label{fig:matching}
  \label{fig:long}
  \label{fig:onecol}
\end{figure*}


Visual localization approaches can be classified into two categories: \textit{Structure-based} and \textit{image-based} methods.  In structure-based methods, the camera pose is estimated from correspondences between 2D points from the query image and a reconstructed 3D point-cloud of the whole scene.  This can lead to great accuracy, but often to mediocre robustness to strong visual changes.  Image-based methods predict the query's camera pose by retrieving the nearest geo-tagged image in a reference database~\cite{Arandjelovic2014DisLocationSD, Chen2011CityscaleLI, DenseVLAD, NetVLAD}.  The advantage is that image retrieval methods can be very robust to strong appearance changes~\cite{DenseVLAD, NetVLAD, ECBR, HFNet}.  The accuracy highly depends on the spatial sampling of the database, but unfortunately high sampling rate is costly both in terms of capture time and memory footprint.  It is therefore natural to combine the two approaches~\cite{Irschara2009FromSP, Middelberg2014Scalable6L, Sarlin2018LeveragingDV, HFNet} into a 'hierarchical' pipeline by finding 2D-3D correspondences only within a subset of the 3D point cloud, obtained using image retrieval.  Such methods benefit from the speed and robustness of image-based approaches, and the accuracy of structure-based methods in lenient capturing conditions.

Still, even when using very recent sparse feature detectors and descriptors~\cite{RootSIFT,Yi16,Ono2018LFNetLL,SuperPoint}, local 2D-3D matching is prone to fail under strong visual changes in practice~\cite{HFNet, 6DOFBenchmark}.  As illustrated in Fig.~\ref{fig:ablation}, it is mostly because it is still difficult to extract the same sparse feature points in two images taken under different conditions.

We therefore propose to detect sparse feature points only in the reference images.  Keeping these sparse feature points is important as they provide the 3D information required to compute the camera pose in an efficient way.  To match these points against the query image, we perform an exhaustive search, which can be implemented efficiently with convolutional operations---the matching procedure takes 10ms on average in our implementation.  Moreover, we notice that the image features extracted by VGG when trained together with NetVLAD to compute a robust global image descriptor provide local descriptions that are remarkably robust to capture condition changes. For our exhaustive search, we therefore rely on these features, which are sometimes called 'Hypercolumns'~\cite{hypercolumns}.

We call the resulting matching method 'Sparse-to-Dense Hypercolumn Matching'. We show that when used together with a powerful retrieval method, it outperforms existing pipelines on several challenging outdoor localization datasets.




The rest of the paper is structured as follows: Section~\ref{relatedwork} discusses the related work while section~\ref{method} introduces our localization pipeline. Our novel `Sparse-to-Dense Hypercolumn Matching' approach is presented in section~\ref{unilateraldetection}. Section~\ref{experiments} describes our experimental setup to thoroughly evaluate our approach in the context of long-term localization, and provides localization results. Source code will be made available.

\begin{figure}[t]
  \begin{center}
    \includegraphics[width=0.95\linewidth]{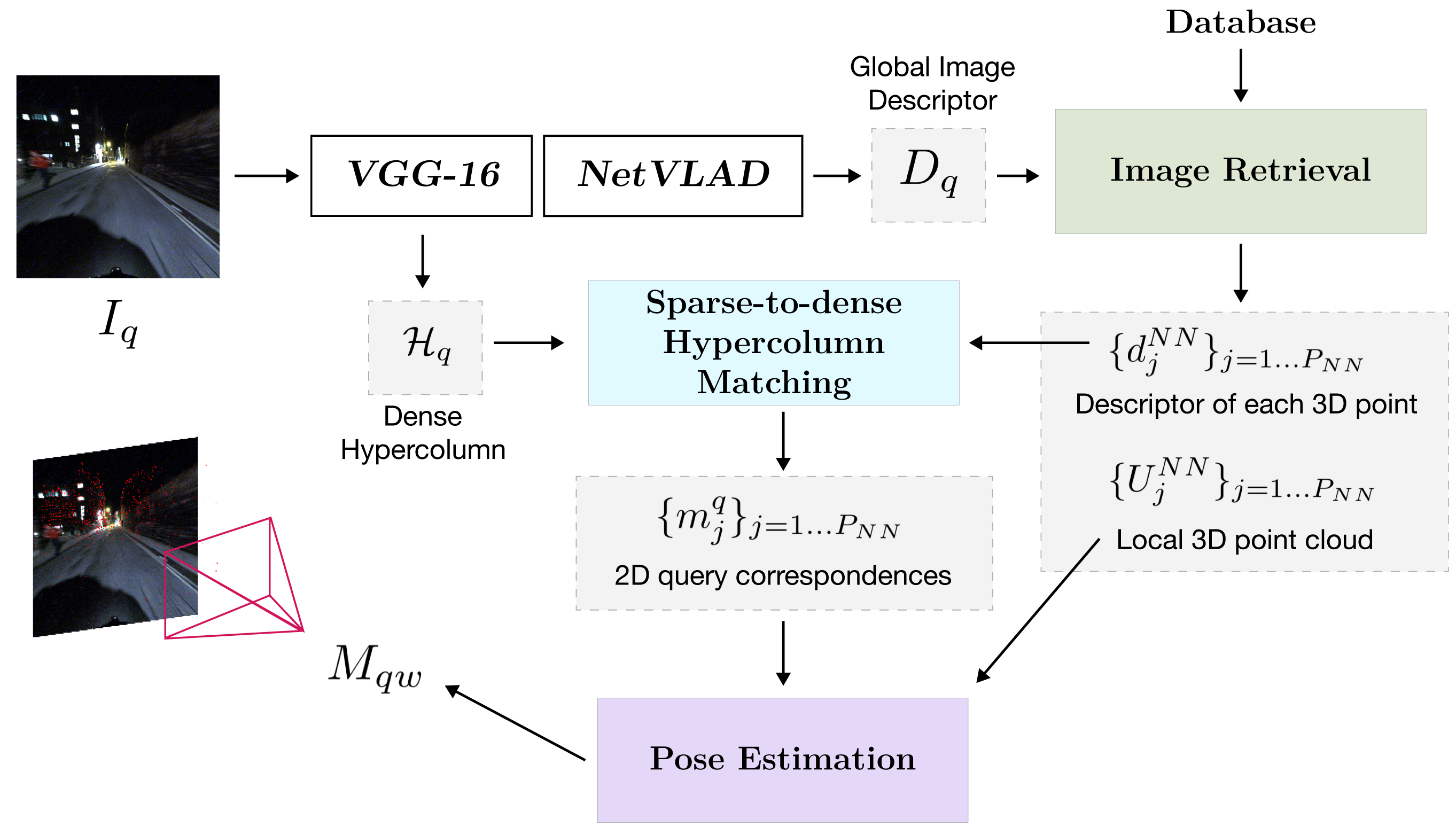}
  \end{center}
  \caption{\textbf{Overview of our hierarchical localization pipeline.}  Given a query image, we compute dense hypercolumns and a global image descriptor using NetVLAD~\cite{NetVLAD}.  The image features are extracted using VGG-16 specifically trained for the image retrieval task under varying capture conditions such as day and night.  We find top-ranked images in a pose-annotated image database, and subsequently use the locally reconstructed point cloud as a feature point detection source.  For each feature, we extract sparse hypercolumns and match each of them exhaustively with the query dense representation.  This results in numerous robust correspondences suitable to perform P$n$P+RANSAC across changing conditions.}\label{fig:method}
\end{figure}

\section{Related Work}\label{relatedwork}

In this section, we review existing approaches tackling the problem of long-term visual localization.  We distinguish \textit{structure-based} methods, which leverage a 3D model of the scene, from \textit{retrieval-based} methods, which do not.

\subsection{Structure-Based Localization}

Structure-based methods regress the full 6~DoF camera pose of query images using direct 2D-3D correspondences. Such methods~\cite{Li2012WorldwidePE,Li:2010:LRU:1888028.1888088,Liu2017EfficientG2,Sattler2015HyperpointsAF,sattler:hal-01513083,CSL} work by first acquiring a point-cloud model of the scene through \textit{SfM}, and computing local feature descriptors like SIFT~\cite{SIFT}, RootSIFT~\cite{RootSIFT} or LIFT~\cite{Yi16}.
These descriptors are in turn used to obtain 2D-to-3D correspondences, and the predicted camera can usually be inferred from those matches using RANSAC~\cite{Fischler1981RandomSC, Sattler2014OnSF} combined with a  Perspective-n-Point~(P$n$P) solver~\cite{Lepetit2008EPnPAA, Bujnak2010NewES, Haralick1994ReviewAA, Kukelova2013RealTimeST}.

In consistent daytime conditions, such methods achieve very competitive results~\cite{Sattler2017EfficientE, CSL, sattler:hal-01513083, Walch2017ImageBasedLU}. However, they rely heavily on the accuracy and robustness of the local 2D-3D correspondences. Research in structure-based approaches mostly focuses on improving descriptor matching efficiency~\cite{Choudhary:2012:VPS:2403138.2403150, Li:2010:LRU:1888028.1888088, Sattler2017EfficientE, Lim2012RealtimeI6, Larsson2016OutlierRF, Lynen2015GetOO}, speed~\cite{Heisterklaus2014ImagebasedPE,Donoser2014DiscriminativeFM} and robustness~\cite{Svrm2014AccurateLA,Sattler2016LargeScaleLR, Li2012WorldwidePE, Sattler2015HyperpointsAF, CSL, Zeisl2015CameraPV}. Yet, under strong condition changes, failures in direct matching start to appear and damage the localization performance~\cite{6DOFBenchmark}. In order to improve the robustness of local feature descriptors and thus increase long-term localization performance, recent methods have used semantic reasoning~\cite{SMC}. Indeed, semantic maps are to some extent condition-invariant, and can enhance either the feature matching stage~\cite{Arandjelovic2014VisualVW,Kobyshev2014MatchingFC,pub.1046137732, Schnberger2017SemanticVL} or the pose estimation stage~\cite{SMC}.
While being accurate at small scale, feature-based methods bottleneck is scalability. In large-scale scenarios, both the construction of precise 3D models~(and their maintenance) and local feature-matching is challenging and expensive~\cite{sattler:hal-01513083}.

\subsection{Image-Based Localization}

In image-based, or retrieval-based, localization methods, accuracy is traded-off for scalability. The scene is modeled as an image database containing ground-truth 6-DoF pose annotations. To infer the pose of a visual query, one can use compact image-level representations to retrieve the top-ranked image from the database and use their labels as pose approximation~\cite{Chen2011CityscaleLI,Zamir2010AccurateIL,Zhang2006ImageBL,sattler:hal-01513083}. The need for ground-truth 3D geometry is alleviated, and this method can easily generalize to large-scale environments.

To obtain robust global image descriptors, one can aggregate local features in the image into a fixed-size representation. VLAD~\cite{Arandjelovic2013AllAV} is a popular descriptor, computed by summing and concatenating many descriptors for affine-invariant regions. DenseVLAD~\cite{DenseVLAD} reformulates the VLAD architecture by densely sampling RootSIFT~\cite{RootSIFT} descriptors in the image. Recent learning-based variants cast the task of image retrieval as a metric learning problem. NetVLAD~\cite{NetVLAD} defines a differentiable VLAD layer as the final activation of a siamese network. Other activations layers~\cite{Tolias2015ParticularOR, Razavian2014VisualIR, Babenko2015AggregatingLD, Kalantidis2016CrossdimensionalWF, Gordo2017EndtoEndLO, GeM} coupled with siamese or triplet architectures, have shown to deliver competitive results for the task of image-retrieval~\cite{Radenovic2018RevisitingOA}.
In a very large database, unsupervised descriptor compression like PCA~\cite{Jgou2012NegativeEA} or Product Quantization~(PQ)~\cite{Jgou2011ProductQF} enables efficient approximate nearest-neighbor search with little loss in performance~\cite{Gordo2017EndtoEndLO}.

Other image-based methods include end-to-end learning approaches, which avoid using explicit feature matching altogether and leverages CNNs to learn robust representations~\cite{Brachmann2017LearningLI,Bui2018SceneCA,Kendall2017GeometricLF, Brachmann2017DSACD}. These methods are either hard to initialize~\cite{6DOFBenchmark, Schnberger2017SemanticVL}, struggle with large environments~\cite{6DOFBenchmark} and/or provide overall poor performance~\cite{Kendall2015PoseNetAC, Brachmann2017LearningLI, Walch2017ImageBasedLU, Balntas_2018_ECCV}.

\subsection{Hierarchical Localization}
For the problem of long-term localization, where strong appearance changes can occur because of the light or season differences, global descriptors have shown to provide robust pose initialization under strong visual changes~\cite{6DOFBenchmark, HFNet, ECBR}. Still, the main bottleneck of retrieval-based localization is the pose approximation step. Several schemes can be implemented to refine the coarsely estimated pose. For instance, view synthesis~\cite{DenseVLAD, Taira2018InLocIV} artificially generates intermediate samples, relative pose regression~\cite{Taira2018InLocIV, Balntas_2018_ECCV} acts as a separate refinement step and multi-image methods~\cite{Zamir2010AccurateIL, Zhang2006ImageBL, Balntas_2018_ECCV} combine the top ranked images to improve pose accuracy.

The image-retrieval step can also be seen as a way to obtain a query's coarse location, before running a structure-based pose refinement algorithm. By doing so, 2D-3D matching is only run on a subset of the whole point cloud, leading to competitive results at small computational costs~\cite{Irschara2009FromSP, Middelberg2014Scalable6L, Sarlin2018LeveragingDV, HFNet}.

\subsection{Learning-Based Feature Matching}

Even in a hierarchical localization pipeline, refining the query camera pose using 2D-3D correspondences can prove to be difficult if the features are not invariant to visual changes and the detections are not consistent across conditions. With the advent of CNNs, learning-based methods for local feature matching have emerged. Methods such as LF-Net~\cite{Ono2018LFNetLL}, SuperPoint~\cite{SuperPoint} or DELF~\cite{Noh2017LargeScaleIR} perform both keypoint detection and feature descriptor computation using end-to-end learning. Under strong condition changes such as day-to-night, even learning-based feature descriptors fail to generalize well~\cite{HFNet}. In this paper, we propose to reuse the pixel-wise dense features directly from the image-retrieval backbone network, and show they are more suited for long-term visual localization.


\section{Method}\label{method}
We give in this section an overview of our pipeline.  We first formalize the problem and its assumption in Section~\ref{problem}. We then provide an overall description of our method in Section~\ref{hierarchical}.


\subsection{Problem Statement}\label{problem}
We assume that a database of registered reference images is available.  More precisely, for each reference image $I_i$ of the database, we assume that the following is available:
\begin{itemize}
  \setlength{\parskip}{0pt}
  \setlength{\itemsep}{0pt plus 1pt}
\item A normalized global image descriptor $D_i$ computed as explained in Section~\ref{hierarchical}, which we will use for the retrieval step.
\end{itemize}
Moreover,
additional information, which we will use for  the pose refinement step, are also stored:
\begin{itemize}
  \setlength{\parskip}{0pt}
  \setlength{\itemsep}{0pt plus 1pt}
\item the calibration matrix $K_i$ and the absolute camera pose $M_{iw}$ expressed in the world coordinate system;
\item a set of $P_i$ 2D feature points $\{m^i_j\}_{j=1...P_i}$ detected using SuperPoint~\cite{SuperPoint}; 
\item the descriptor $d^i_j$ for each feature point $m_j^i$ computed as explained in Section~\ref{unilateraldetection};
\item the 3D coordinates $U^i_j$ of each feature point $m_j^i$.
\end{itemize}


Given a  query image $I_q$ with known calibration matrix $K_q$, and this database, we aim to predict the camera pose $M_{qw}$.

\subsection{Our Hierarchical Localization Pipeline}\label{hierarchical}
When performing localization in large-scale environments, matching a set of 2D keypoints with a large number of 3D landmarks can be difficult~\cite{sattler:hal-01513083}. As suggested by~\cite{HFNet}, one way to reduce the set of 3D points to match the image keypoints against is to first perform image retrieval. The returned top-ranked images in the database provide us with a subset of the large 3D point cloud for which performing local feature matching is much more efficient. The whole pipeline is presented in Figure~\ref{fig:method}.

\paragraph{Image Retrieval.} Like previous methods~\cite{Babenko2015AggregatingLD,                  Tolias2015ParticularOR,  Razavian2014VisualIR, NetVLAD, Kalantidis2016CrossdimensionalWF, Gordo2017EndtoEndLO,GeM}, we use a Siamese network approach to learn a discriminative image descriptor robust to changes of the capture conditions.  For the architecture, we opt for the popular NetVLAD~\cite{NetVLAD} pooling layer with a VGG-16~\cite{VGG16} backbone.



During training, we define positive and negative labels $l(I_i,I_j)\in \{0,1\}$ for pairs of images, based on the presence or absence of co-visibility between images respectively.  We use the same contrastive loss as \cite{NetVLAD}. Once trained, the network provides a global descriptor $D_i$ for each reference image, which is stored in the database.

At test time, given a query image $I_q$, we compute its descriptor $D_q$ and retrieve its $k$ nearest neighbors by computing the Euclidean distance between $D_q$ and each stored descriptor $D_i$.  Such top-ranked images provide coarse camera poses which are sufficient to estimate a query's emplacement~\cite{6DOFBenchmark}.


\paragraph{Camera Pose Refinement.} In order to obtain a more accurate camera pose estimation, we make use of the local 3D point clouds fetched from the image retrieval step. For each of the $k$ nearest neighbors, we establish 2D-3D correspondences and subsequently solve the pose using for instance a Perspective-n-Point (P$n$P)~\cite{Bujnak2010NewES, Haralick1994ReviewAA, Kukelova2013RealTimeST} solver. Given a set of matches, we refine the query pose using P3P~\cite{P3P} inside a RANSAC~\cite{RANSAC, Sattler2014OnSF} loop. The method we use to establish these correspondences is our main contribution, and we describe  it below.

\section{Sparse-to-Dense Hypercolumn Matching}
\label{unilateraldetection}

If we followed the standard approach to obtain the 2D-3D correspondences needed to estimate the camera pose, we would extract sparse feature points in the query image and match them against the sparse feature points $m_j^i$ extracted from the nearest neighbors of the query image. As mentioned in the introduction,
this step is still very challenging, mostly because of the detection step that needs to identify the same image locations even under strong condition changes. In order to circumvent this challenging detection problem, we reformulate the local feature matching step to avoid performing detection in the query image, as illustrated in Fig.~\ref{fig:matching}. To do so, we perform an exhaustive search in the query image for the correspondent of each sparse feature point detected in the reference images. We explain below how this search can be performed efficiently.


%



\paragraph{HyperColumn Extraction.}
In order to perform robust matching, we rely on image features that were already used to compute the global image descriptor as shown in Fig.~\ref{fig:method}.
For each query image, we extract intermediate features from the VGG-16~\cite{VGG16} network and aggregate them in order to obtain a dense and rich representation of the image. \hugo{We extract features from the layers \code{conv\_3\_3}, \code{conv\_4\_1}, \code{conv\_4\_3}, \code{conv\_5\_1}, \code{conv\_5\_3}}. We refer to these representations as ``hypercolumns''~\cite{hypercolumns}. Each intermediate layer is  upsampled using bilinear interpolation to match the resolution $W_{\mathcal{H}} \times H_{\mathcal{H}}$ of the earliest layer, before being concatenated along the channel axis and normalized. We define the obtained hypercolumns for the query image $I_q$ as $\mathcal{H}_q\in\mathbb{R}^{W_{\mathcal{H}} \times H_{\mathcal{H}} \times C}$.

For each reference image $I_i$, we are only interested in descriptors located at feature points. We thus only store in the database the hypercolumns at locations $\{m^i_j\}_{j=1...P_i}$.  We denote $\mathcal{S}_i = \{d_j^i\}_{j=1..P_i}$ this set of sparse descriptors, where $d_j^i\in\mathbb{R}^{1 \times 1 \times C}$.\\

\begin{figure*}[t]
  \begin{center}
    \includegraphics[width=0.95\linewidth]{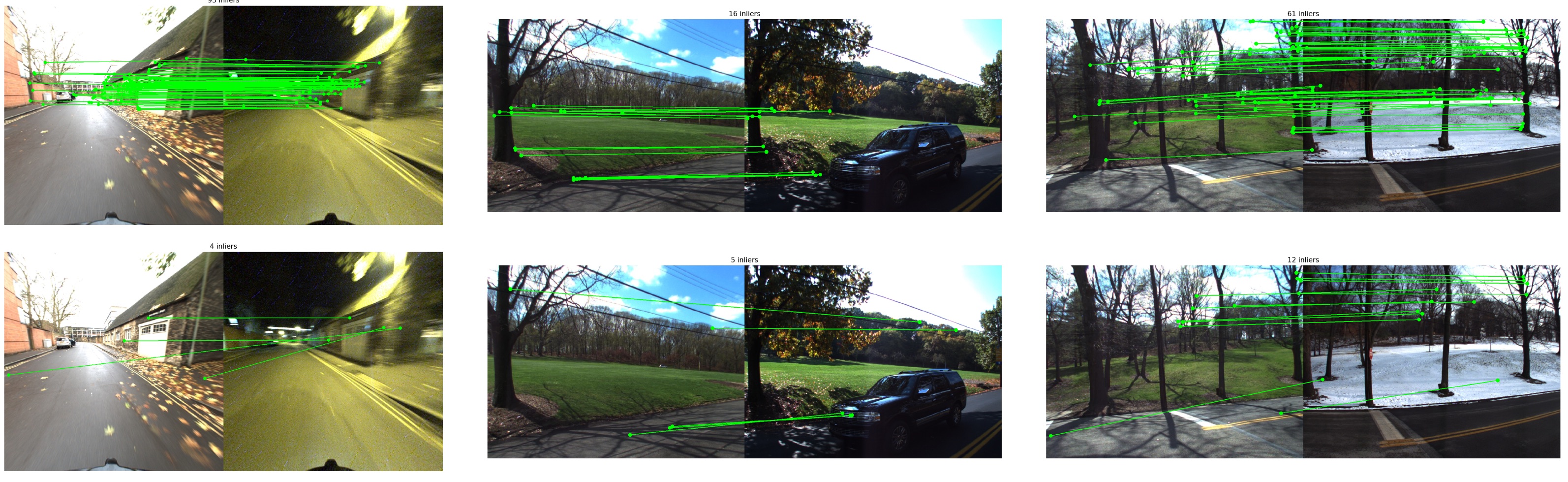}
  \end{center}
  \vspace{-4.5mm}
  \caption{\textbf{Examples of inlier correspondences obtained using RANSAC+PnP.}  Top-row shows correspondences obtained with our \commented{`Sparse-to-Dense Hypercolumn Matching'} method, bottom row shows correspondences obtained with SuperPoint detection and descriptors.}\label{fig:exemples_corresp}
\end{figure*}
\begin{figure}[t]
  \begin{center}
    \includegraphics[width=0.95\linewidth]{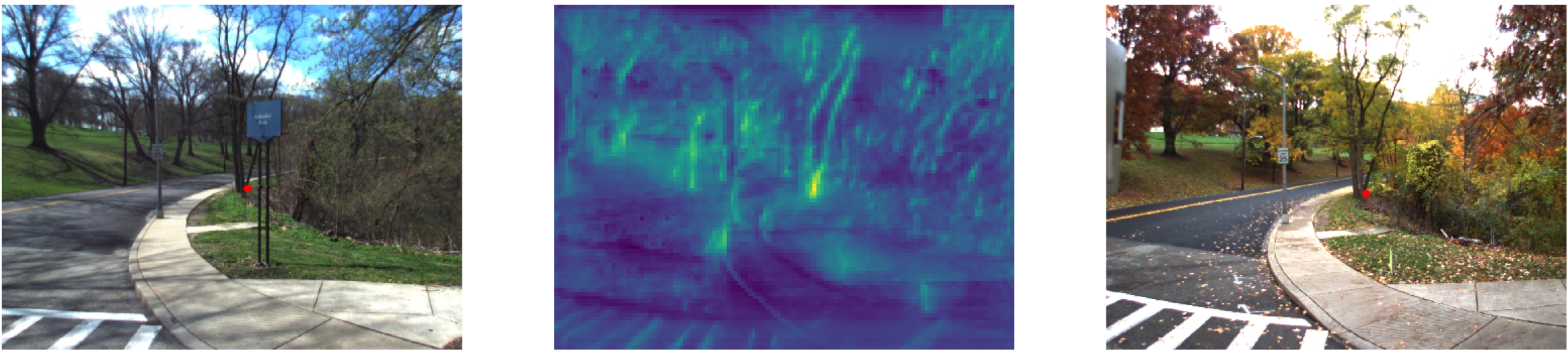}
  \end{center}
  \vspace{-4mm}
  \caption{\textbf{Example of correlation map.}  Left image shows a Superpoint in the reference image. The corresponding sparse hypercolumn descriptor is used to compute the correlation map (middle) and retrieve the 2D correspondent in the query image (right).}\label{fig:heatmap} 
\end{figure}

\noindent\textbf{Sparse-to-Dense Matching.}
To find correspondences between the set of sparse descriptors from the reference image $\mathcal{S}_i$ and the dense hypercolumns $\mathcal{H}_q$, we perform a dot product. These dot products can be efficiently implemented with a 1$\times$1 convolution. We define the resulting cross-correlation map as $\mathcal{C}_{q,j}^i = \mathcal{H}_q * d_j^i\in\mathbb{R}^{W_{\mathcal{H}} \times H_{\mathcal{H}}}$.
%
To retrieve the final 2D keypoints in the query image, we first fetch the global maximum of the cross-correlation map and upsample the retrieved coordinates to match the query image coordinates. Consequently, this `Sparse-to-Dense matching' step always gives us $P_i$ 2D-3D correspondences (See Figures~\ref{fig:exemples_corresp} and \ref{fig:heatmap}).\\




\noindent\textbf{Ratio Test.}
Some detections in the reference image may fall in image regions with repetitive textures, or in areas that are occluded in the query image. This may lead to ambiguities when looking for point correspondents. To discard matches with large ambiguity, we apply a ratio test similar to the one often used in more standard approaches, and defined as follows.
For the cross-correlation map $\mathcal{C}_{q,j}^{i}$, let $\bar{\mathcal{C}}_{q,j}^{i}\in\mathbb{R}^{(W_{\mathcal{H}} . H_{\mathcal{H}})}$ be the flattened and sorted by decreasing order map. For a 2D-3D match to be retained, we apply the following rule:
\begin{equation}\frac{\bar{\mathcal{C}}_{q,j}^{i}[0]}{\bar{\mathcal{C}}_{q,j}^{i}[f \times (W_{\mathcal{H}} \times H_{\mathcal{H}})]} > \alpha , f\in[0;1] \> .\end{equation}
In practice, we use $\alpha = 0.9$, and adapt the factor $f$ to the different datasets. Finding the
value of $\bar{\mathcal{C}}_{q,j}^{i}[f \times (W_{\mathcal{H}} \times H_{\mathcal{H}})]$ actually does not require sorting the whole array,  and adds negligible overload to the computational cost.

\section{Experiments}\label{experiments}
In this section, we conduct experiments to evaluate our hierarchical localization approach under challenging conditions. In Section~\ref{evaluation}, we detail how both our evaluation datasets were setup and reconstructed. We also discuss the evaluation methods and baselines used for comparison. In Section~\ref{localization}, we show how our hierarchical method can solve camera poses accurately under challenging conditions and outperforms existing methods in such categories. Lastly, in Section~\ref{ablation}, we run an ablation study, which demonstrates the improvements brought by our contribution.

\begin{table}
  \ra{1.0}
  \begin{center}
    \resizebox{\columnwidth}{!}{%
      \begin{tabular}{@{}cccccc@{}}
        \toprule
        Dataset & \makecell{Training\\sequences} & Condition & \makecell{Training\\images} &  \makecell{Reference\\images} & \makecell{Query\\images} \\
        \midrule
        \multirow{4}{*}{\makecell{RobotCar\\Seasons\\\cite{RobotCar}}}
        & 12 Dec 2014 & overcast & 20,965 & \multirow{4}{*}{6,954} & \multirow{4}{*}{3,978}\\
        & 05 Dec 2014 & overcast-rain & 20,965 &\\
        & 16 Dec 2014 & night & 19,376 &\\
        & 03 Feb 2015 & night & 20,257 &\\
        \hline
        \multirow{3}{*}{\makecell{Extended\\CMU-Seasons\\\cite{CMUSeasons}}}
        & Slices 2-8 & urban & 9,612 & \multirow{3}{*}{7,159} & \multirow{3}{*}{75,335}\\
        & Slices 9-17 & suburban & 24,728 &\\
        & Slices 18-25 & park & 16,148 &\\
        \hline

        \bottomrule
      \end{tabular}%
    }
  \end{center}
 \vspace{-3mm}
  \caption{\textbf{Detailed statistics} regarding the training and testing sequences used for each dataset. Reference images are used to triangulate 3D keypoints offline using SuperPoint~\cite{SuperPoint} detections and descriptors. Note that for RobotCar Seasons, only rear images are considered.}\label{table:datasets}
\end{table}

\subsection{Evaluation Setup}\label{evaluation}
We begin our evaluation by presenting the two challenging outdoor datasets introduced by~\cite{6DOFBenchmark} which we will be using throughout this section.\\

\noindent\textbf{Datasets.} Our evaluation set consists of two outdoor datasets captured from vehicles or using hand-held mobile phone cameras. Each of the provided datasets contains a set of reference images, along with their ground truth camera poses. We are also given sparse 3D reconstructions pre-computed using RootSIFT~\cite{RootSIFT} features by Sattler~\etal\cite{6DOFBenchmark}. In practice, we do not use the provided sparse 3D reconstruction and re-triangulated our own point clouds using SuperPoint~\cite{SuperPoint} detections. We perform the triangulation using COLMAP~\cite{COLMAP1, COLMAP2} on the reference images of each dataset, similarly to~\cite{HFNet}.



The first dataset is the Extended CMU-Seasons dataset~\cite{6DOFBenchmark}, which contains about 40\% more images than the original CMU-Seasons dataset~\cite{CMUSeasons}. It consists of 7,159 reference images and 75,335 query images, captured using two front-facing cameras mounted on a car, in the area of Pittsburgh. The images were captured over the course of a year and the reference images depict different seasonal conditions. The \textit{park} scene is particularly difficult as it was captured in a rural environment and faces strong vegetation changes over the year.

The second dataset is the RobotCar Seasons dataset~\cite{RobotCar}, which contains 6,954 daytime images captured by a rear-facing camera mounted on a car driving in Oxford. The 3,978 query images were taken over the course of a year, including some in very challenging conditions such at nighttime~\cite{6DOFBenchmark}. Note that in this paper we do not consider the additional reference images taken by the two side-facing cameras. We report details about the exact sequences used for training for each dataset in Table~\ref{table:datasets}.\\

\begin{table*}
  \ra{1.05}
  \begin{center}
    \resizebox{\textwidth}{!}{
      \begin{tabular}{@{} ll *{17}{c@{\hskip0.1in}} @{}}%
        & \multicolumn{1}{c}{}
        & \multicolumn{6}{c}{\textit{RobotCar Seasons}}
        & \multicolumn{9}{c}{\textit{Extended CMU-Seasons}}\\
        \cmidrule(lr){3-8}
        \cmidrule(lr){9-17}
        \multicolumn{2}{c}{}
        & \multicolumn{3}{c}{\textit{Day-All}}
        & \multicolumn{3}{c}{\textit{Night-All}}
        & \multicolumn{3}{c}{\textit{Urban}}
        & \multicolumn{3}{c}{\textit{Suburban}}
        & \multicolumn{3}{c}{\textit{Park}}\\
        \toprule
        \multicolumn{2}{c}{Method}
        & \multicolumn{3}{c}{Threshold Accuracy}
        & \multicolumn{3}{c}{Threshold Accuracy}
        & \multicolumn{3}{c}{Threshold Accuracy}
        & \multicolumn{3}{c}{Threshold Accuracy}
        & \multicolumn{3}{c}{Threshold Accuracy}\\
        \cmidrule(lr){3-5} \cmidrule(lr){6-8}
        \cmidrule(lr){9-11} \cmidrule(lr){12-14} \cmidrule(lr){15-17} &
        &\makecell{0.25m \\ 2\degree} & \makecell{0.5m \\ 5\degree} & \makecell{5m \\ 10\degree}
        &\makecell{0.25m \\ 2\degree} & \makecell{0.5m \\ 5\degree} & \makecell{5m \\ 10\degree}
        &\makecell{0.25m \\ 2\degree} & \makecell{0.5m \\ 5\degree} & \makecell{5m \\ 10\degree}
        &\makecell{0.25m \\ 2\degree} & \makecell{0.5m \\ 5\degree} & \makecell{5m \\ 10\degree}
        &\makecell{0.25m \\ 2\degree} & \makecell{0.5m \\ 5\degree} & \makecell{5m \\ 10\degree}\\
        \midrule
        \parbox[t]{6mm}{\multirow{3}{*}{\rotatebox[origin=c]{90}{\makecell{Structure\\-based}}}} &
        CSL~\cite{CSL} & 45.3 & 73.5 & 90.1 & 0.6 & 2.6 & 7.2
                        & 71.2 & 74.6 & 78.7 & 57.8 & 61.7 & 67.5 & 34.5 & 37.0 & 42.2\\
        & AS~\cite{ActiveSearch} & 35.6 & 67.9 & 90.4 & 0.9 & 2.1 & 4.3
                       & - & - & - & - & - & - & - & - & -\\
        & SMC~\cite{SMC} & \boldblue{50.3} & \boldred{79.3} & \boldred{95.2} & \boldblue{7.1} & \boldblue{22.4} & 45.3
                       & \boldblue{88.8} & \boldblue{93.6} & \boldblue{96.3}	& \boldred{78.0} & \boldred{83.8} & 89.2 & \boldred{63.6} & \boldblue{70.3} & 77.3\\
        \hline
        \parbox[t]{6mm}{\multirow{4}{*}{\rotatebox[origin=c]{90}{\makecell{Retrieval\\-based}}}} &
        FAB-MAP~\cite{Cummins2008FABMAPPL} & 2.7 & 11.8 & 37.3 & 0.0 & 0.0 & 0.0
                                            & - & - & - & - & - & - & - & - & -\\
        & NetVLAD~\cite{NetVLAD} & 6.4 & 26.3 & 90.9 & 0.3 & 2.3 & 15.9
                                            & 12.2 & 31.5 & 89.8 & 3.7 & 13.9 & 74.7 & 2.6 & 10.4 & 55.9\\
        & DenseVLAD~\cite{DenseVLAD} & 7.6 & 31.2 & 91.2 & 1.0 & 4.4 & 22.7
                                     & 14.7 & 36.3 & 83.9 & 5.3 & 18.7 & 73.9&5.2 & 19.1 & 62.0\\
        & ToDayGAN~\cite{ToDayGAN} & 7.6 & 31.2 & 91.2 & 2.2 & 10.8 &\boldblue{50.5}
                                     & - & - & - & - & - & - & - & - & -\\
        \hline
        \parbox[t]{6mm}{\multirow{2}{*}{\rotatebox[origin=c]{90}{\makecell{Hierar\\-chical}}}} &
        NV+SP~\cite{HFNet} & \boldred{53.0} & \boldred{79.3} & 95.0 & 5.9 & 17.1 & 29.4
                          & \boldred{89.5} & \boldred{94.2} & \boldred{97.9} & \boldblue{76.5} & \boldblue{82.7} & \boldblue{92.7} & \boldblue{57.4} & 64.4 & \boldblue{80.4}\\
        & \textbf{NV-r + S-D + H (Ours)} & 45.7 & \boldblue{78.0} & \boldblue{95.1} & \boldred{22.3} & \boldred{61.8} & \boldred{94.5}
                          & 65.7 & 82.7 & 91.0 & 66.5 & 82.6 & \boldred{92.9} & 54.3 & \boldred{71.6} & \boldred{84.1}
        \\
        \bottomrule
      \end{tabular}%
    }
  \end{center}
      \vspace{-2.5mm}
  \caption{\textbf{Localization results}. We report localization
    recalls in percent, for three translation and orientation
    thresholds (\textit{high}, \textit{medium},  and
    \textit{coarse}) as in \cite{6DOFBenchmark}. We highlight the \boldred{best} in red and
    \boldblue{second-best} in  blue performances for each
    threshold. Note that NetVLAD, ToDayGAN, and NV+SP all use
    pre-trained NetVLAD weights from Pittsburgh30k~\cite{NetVLAD},
    while we retrained ours on other RobotCar sequences. We also
    include SMC, which uses additional semantic data and
    assumptions. For Extended CMU-Seasons, some methods did not
    provide results for the benchmark.}\label{table:results}
\end{table*}

\noindent\textbf{Baselines.} We compare our approach both against structure-based and retrieval-based state-of-the-art methods. Localization results for these methods were provided by the authors of the benchmark~\cite{6DOFBenchmark}.

For structure-based methods, we compare our approach to Active Search (AS)~\cite{ActiveSearch} and City-Scale Localization~(CSL)~\cite{CSL}. Both methods are direct 2D-3D matching techniques optimized for matching efficiency and robustness respectively, and have shown to deliver great accuracy in daytime conditions at a high precision threshold~\cite{6DOFBenchmark}. We also display results for Semantic Match Consistency~(SMC)~\cite{SMC}, which leverages semantic maps to filter outliers in the matching stage, and makes additional assumptions regarding the camera height and gravity vector.

We also compare our approach to retrieval-based methods, such as NetVLAD (pre-trained on Pittsburgh30k~\cite{NetVLAD} with a VGG-16~\cite{VGG16} backbone), and to DenseVLAD~\cite{DenseVLAD}. For these methods, we simply approximate the query image camera pose by the pose of its retrieved top-ranked database image. Details about their configuration and implementation details can be found in the original benchmark~\cite{6DOFBenchmark}. Additionally for RobotCar Seasons, we report the results obtained by performing night-to-day image translation using a GAN architecture (ToDayGAN)~\cite{ToDayGAN}, prior to running DenseVLAD. Lastly, we show the results obtained by Sarlin~\etal~\cite{HFNet}, which is a hierachical approach using a pre-trained NetVLAD backbone followed by SuperPoint~\cite{SuperPoint} feature detection and local descriptors for 2D-3D matching (NV+SP). This method also uses co-visibility clusters to merge 3D points from neighbouring database images.

\paragraph{Metrics.} We evaluate our approach using the same localization metric as~\cite{6DOFBenchmark}. Three precision thresholds are defined, accounting for both positional and rotational error. We refer to these thresholds as \textit{high} (0.25m and 2\degree), \textit{medium} (0.5m and 5\degree) and \textit{coarse} (5m and 10\degree) precision. For each threshold, we report the localization recall in percent.

\subsection{Large-Scale Localization}\label{localization}
Having established our evaluation process, we now report the performance of our approach.\\

\noindent\textbf{Training sets.}
For the NetVLAD retrieval backbone, we use different weights for both datasets. For RobotCar Seasons~\cite{RobotCar}, we retrained NetVLAD on tuples extracted from other RobotCar sequences, featuring for daytime and nighttime images (see Table~\ref{table:datasets}). Positive and negative tuples were assembled using the provided GPS and INS data. Note that these sequences do not overlap with the test set. For Extended CMU-Seasons~\cite{CMUSeasons}, we built training samples using all the provided annotated training data from the \textit{urban}, \textit{suburban} and \textit{park} slices. 
When training NetVLAD, we use hard-negative mining at every epoch, to obtain for each query the hardest subset of all possible negatives in the database.\\

\noindent\textbf{Methods.} 
As presented in Section~\ref{method}, we run our hierarchical localization pipeline by first ranking each query with respect to the reference images. We use the normalized global image descriptors produced by NetVLAD~(NV), and obtain the rankings using a simple dot product. To account for potential image retrieval errors, for every query we run the exhaustive matching step on each of the top-$N$ nearest neighbors. The final predicted pose is picked as the one having the highest number of inliers in the RANSAC loop of the PnP. For RobotCar Seasons, we use $N=15$ and for Extended CMU-Seasons, we use $N=10$ because of the large amount of images to evaluate.


\begin{table}
  \ra{1.0}
  \begin{center}
    \resizebox{\columnwidth}{!}{
      \begin{tabular}{@{} l *{6}{c@{\hskip0.1in}} @{}}%
        \multicolumn{1}{c}{}
        & \multicolumn{3}{c}{\textit{Day-All}}
        & \multicolumn{3}{c}{\textit{Night-All}}\\
        \toprule
        \multicolumn{1}{c}{Method}
        & \multicolumn{3}{c}{Threshold Accuracy}
        & \multicolumn{3}{c}{Threshold Accuracy}\\
        \cmidrule(lr){2-4} \cmidrule(lr){5-7}&
        \makecell{0.25m \\ 2\degree} & \makecell{0.5m \\ 5\degree} & \makecell{5m \\ 10\degree}
        &\makecell{0.25m \\ 2\degree} & \makecell{0.5m \\ 5\degree} & \makecell{5m \\ 10\degree}\\
        \midrule
        NV (pre-trained) & 6.4 & 26.3 & 90.9 & 0.3 & 2.3 & 15.9\\
        NV-r (re-trained) & 4.1 & 17.8 & 86.9 & 2.4 & 11.4 & 84.6\\
        NV-r + S-S + SP & \textbf{52.9} & \textbf{78.5} & 93.8 & 10.9 & 32.7 & 87.4\\
        NV-r + S-S + H & 49.0 & 77.9 & 93.6 & 14.8 & 44.5 & 89.7\\
        NV-r + S-D + SP & 50.3 & 77.5 & 92.9 & 14.4 & 43.2 & 87.8\\
        NV-r + S-D + H & 45.7 & 78.0 & \textbf{95.1} & \textbf{22.3} & \textbf{61.8} & \textbf{94.5}\\
        \bottomrule
      \end{tabular}%
    }
  \end{center}
   \vspace{-2.5mm}
  \caption{\textbf{Ablation Study} on the RobotCar Seasons dataset. We first show the improvements coming from using a retrained NetVLAD~(NV)~\cite{NetVLAD} backbone. Then, we report localization performance using standard `Sparse-to-Sparse' (S-S) matching using SuperPoint detections and two different descriptors: SuperPoint descriptors~(S-S + SP) and Hypercolumn descriptors (S-S + H), as well as the results of our 'Sparse-to-dense' (S-D) matching using SuperPoint descriptors~(S-D + SP) and Hypercolumn descriptors~(S-D + H). We report localization recall in percent, for three translation and orientation thresholds.}\label{table:ablation}
\end{table}

\begin{table}
  \ra{1.0}
  \begin{center}
    \resizebox{\columnwidth}{!}{
      \begin{tabular}{@{} l *{5}{c@{\hskip0.1in}} @{}}%
        \toprule
         & \makecell{Dense Query\\Hypercolumn\\Descriptors}
         & \makecell{\textit{Sparse Reference} \\ \textit{Hypercolumn}\\ \textit{Descriptors} \\ \textit{(offline)}}
         & \makecell{\textit{Correspondence Maps}\\ \textit{(Exhaustive search)}}
         & \makecell{\textit{Ratio Test}\\\textit{(non-optimized)}}
         & \makecell{\textit{PnP Solving}}\\
        \midrule
        Runtime (ms) & 107.29 &  114.71 & 10.8 & 169.14 & 3.08 \\ 
        \bottomrule
      \end{tabular}%
    }
  \end{center}
    \vspace{-3mm}
  \caption{\textbf{Runtime measurements.} We report the average runtimes for our sparse-to-dense matching approach on RobotCar Season, with $512 \times 512$ input images. Operations in italic are run for each of the top-ranked images.}\label{table:runtimes}
\end{table}

\paragraph{Implementation details.} We use a Pytorch implementation of NetVLAD to compute the global image descriptors as well as the intermediate VGG-16 features used to compute the hypercolumns. As in~\cite{HFNet}, we reduce the dimensionality of all produced descriptors to a size of 1024 using PCA, learned on the reference set. When retraining NetVLAD on RobotCar Seasons and Extended CMU-Seasons, images are rescaled to a maximum size of 512 pixels, while preserving image ratio. At inference time, we again rescale images to a maximum size of 512 pixels for all datasets, both to compute the global image descriptors and to extract intermediate dense features. The offline point cloud triangulation and the online 2D-3D correspondences are done using the original images resolutions.

We use different ratio test values for each dataset. For RobotCar Seasons we use a factor of $f=0.006$. For Extended CMU-Seasons we use a value of $0.12$, as we found much more ambiguous matches and using selective thresholds were leading to a high number of rejections.
As in~\cite{HFNet}, for both datasets, the RANSAC~\cite{RANSAC} loop stops
when a pose has a minimum number of inliers of 15.
\paragraph{Performance.}
We run our experiments on a PC equipped with an Intel(R) Xeon(R) E5-2630 CPU (2.20GHz) CPU with 128GB of RAM and an NVIDIA GeForce GTX 1080Ti GPU.  \hugo{We pre-compute compressed global image descriptors for a faster image retrieval at inference time.  \vincent{Our main bottleneck in terms of computation times} in our current implementation lies in the VGG-16 inference.  As shown in~\cite{HFNet}, this part can be sped up using a teacher network with little loss in accuracy. Our ratio test method could also be replaced by a faster, more traditional non-maxima suppression scheme computed on GPU.} The computation of the correspondence map is done on GPU through a convolution operation and takes on average 30ms in our implementation (depending on the input image resolution and ratio).  We report the average measured runtimes in Table~\ref{table:runtimes}.

\paragraph{Results.} We report the localization results in Table~\ref{table:results}. Our method outperforms all baselines in very challenging scenarios such as nighttime for RobotCar Seasons. We also show significant improvements for the \textit{park} scene of Extended CMU-Seasons, which is arguably the most difficult with strong changes in vegetation, at \textit{medium} and \textit{coarse} precision thresholds. For other categories, the performance is usually on par with state-of-the-art structure-based or hierarchical methods such as SMC~\cite{SMC} or NV+SP~\cite{HFNet} respectively.
\hugo{\vincent{On} easier categories, such as \textit{day-all} for RobotCar Seasons or \textit{urban} for CMU,
  our approach is not as accurate as other feature-point based approaches, especially at a finer threshold.
It is therefore more adapted to complex correspondence problems. On less challenging cases, the standard approach which relies on a detector with sub-pixel accuracy for the query image can still be more accurate.}


\subsection{Ablation Study}\label{ablation}
Having presented the results of our full pipeline, we now evaluate the impact of each element of our pipeline in the localization step. We run this ablation study on RobotCar Season~\cite{RobotCar} and report our results in Table~\ref{table:ablation}.\\

\noindent\textbf{NetVLAD backbones.} We first discuss the impact of having a retrained
image-retrieval backbone. As shown in Table~\ref{table:ablation}, the pre-trained Pittsburgh30k~\cite{NetVLAD} weights (NV) provides a good coarse pose estimation in daytime, but still very mild results at nighttime. We can already see that this will be a very limiting factor when performing 2D-3D matching, as the selected point cloud subsets will not be overlapping with the query image. When retraining NetVLAD (NV-r) on nighttime sequences from RobotCar, this gives a significant boost in performance, especially at a coarse precision level.

However, this is also tightly linked with the database spatial sampling: A dataset sampled much more sparsely would yield poor results at a coarse level even at daytime. We also tried retraining NetVLAD with a ResNet-50~\cite{ResNet} backbone, and / or a GeM~\cite{GeM} layer activation, but this always yielded slightly poorer retrieval results than a VGG-16~\cite{VGG16} network with a VLAD activation layer.\\

\noindent\textbf{`Sparse-to-Sparse' Matching.} We evaluate adding a subsequent camera pose estimation using 2D-3D matches coming from standard `Sparse-to-Sparse' (S-S) matching using SuperPoint~\cite{SuperPoint} detections and two different descriptors: SuperPoint descriptors (S-S + SP) and Hypercolumn descriptors (S-S + H). Both approaches (S-S + SP) and (S-S + H) allow to significantly improve the daytime results. For nighttime results, even if the performance improved, they remain limited compared to the daytime. We argue that this discrepancy between daytime and nighttime results comes from the difficulty to detect and match sparse feature points extracted from two images captured under very different conditions. This motivates our novel `Sparse-to-Dense' matching approach. Finally, one can see that the aggregation of dense features into hypercolumns at different levels provides improvements. This shows the advantage of using hypercolumns for description rather than the Superpoint descriptors. This advantage is likely due to the large receptive fields of the hypercolumns computed  by VGG, and the way they are learned to be condition-invariant.\\

\noindent\textbf{`Sparse-to-Dense' Matching.} We finally evaluate replacing the standard `Sparse-to-Sparse' matching with our novel `Sparse-to-Dense' matching for both Superpoint descriptors (S-D + SP) and Hypercolumn descriptors (S-D + H). As shown in Table~\ref{table:ablation}, our novel approach is a way to partially remove the nighttime detection bottleneck: Compared to `Sparse-to-Sparse' Hypercolumn matching (NV-r + S-S + H), our `Sparse-to-Dense' Hypercolumn matching (NV-R + S-D + H) increases the recall by $7.5\%$ and $17.3\%$ for the \textit{high} and \textit{medium} thresholds respectively at nighttime.






\section{Conclusion}
We have introduced a novel hierarchical localization method, which reformulates the 2D-3D matching stage to improve long-term localization capabilities. We showed that  breaking the paradigm of detecting feature points in both images to match, we can significantly improve the number of correct matches. While this approach was demonstrated in this paper in the context of localization, it is very likely to be useful for other applications.



\section*{Acknowledgement}
This project has received funding from the Bosch Research Foundation~(\emph{Bosch Forschungsstiftung}). We gratefully acknowledge the support of NVIDIA Corporation with the donation of the Titan Xp GPU used for this research. The authors would also like to warmly thank the authors of the benchmark~\cite{6DOFBenchmark} for providing support with the evaluation tools. Vincent Lepetit is a senior member of the \emph{Institut Universitaire de France}~(IUF).


{\small
  \bibliographystyle{ieee}
  \bibliography{string,cleaned_refs2}

\begin{thebibliography}{10}\itemsep=-1pt

\bibitem{ToDayGAN}
A.~Anoosheh, T.~Sattler, R.~Timofte, M.~Pollefeys, and L.~{Van~Gool}.
\newblock {Night-To-Day Image Translation for Retrieval-Based Localization}.
\newblock {\em CoRR}, abs/1809.09767, 2018.

\bibitem{NetVLAD}
R.~Arandjelovic, P.~Gron{\'a}t, A.~Torii, T.~Pajdla, and J.~Sivic.
\newblock {NetVLAD: CNN Architecture for Weakly Supervised Place Recognition}.
\newblock In {\em Conference on Computer Vision and Pattern Recognition}, pages
  5297--5307, 2016.

\bibitem{RootSIFT}
R.~Arandjelovic and A.~Zisserman.
\newblock {Three Things Everyone Should Know to Improve Object Retrieval}.
\newblock In {\em Conference on Computer Vision and Pattern Recognition}, pages
  2911--2918, 2012.

\bibitem{Arandjelovic2013AllAV}
R.~Arandjelovic and A.~Zisserman.
\newblock {All About VLAD}.
\newblock In {\em Conference on Computer Vision and Pattern Recognition}, pages
  1578--1585, 2013.

\bibitem{Arandjelovic2014DisLocationSD}
R.~Arandjelovic and A.~Zisserman.
\newblock {Dislocation: Scalable Descriptor Distinctiveness for Location
  Recognition}.
\newblock In {\em Asian Conference on Computer Vision}, 2014.

\bibitem{Arandjelovic2014VisualVW}
R.~Arandjelovic and A.~Zisserman.
\newblock {Visual Vocabulary with a Semantic Twist}.
\newblock In {\em Asian Conference on Computer Vision}, 2014.

\bibitem{Babenko2015AggregatingLD}
A.~Babenko and V.~S. Lempitsky.
\newblock {Aggregating Local Deep Features for Image Retrieval}.
\newblock In {\em International Conference on Computer Vision}, pages
  1269--1277, 2015.

\bibitem{CMUSeasons}
H.~Badino, D.~F. Huber, and T.~Kanade.
\newblock {Visual Topometric Localization}.
\newblock {\em 2011 IEEE Intelligent Vehicles Symposium (IV)}, pages 794--799,
  2011.

\bibitem{Balntas_2018_ECCV}
V.~Balntas, S.~Li, and V.~Prisacariu.
\newblock {Relocnet: Continuous Metric Learning Relocalisation Using Neural
  Nets}.
\newblock In {\em European Conference on Computer Vision}, 09 2018.

\bibitem{Brachmann2017DSACD}
E.~Brachmann, A.~Krull, S.~Nowozin, J.~Shotton, F.~Michel, S.~Gumhold, and
  C.~Rother.
\newblock {DSAC — Differentiable RANSAC for Camera Localization}.
\newblock In {\em Conference on Computer Vision and Pattern Recognition}, pages
  2492--2500, 2017.

\bibitem{Brachmann2017LearningLI}
E.~Brachmann and C.~Rother.
\newblock {Learning Less is More - 6D Camera Localization via 3D Surface
  Regression}.
\newblock {\em CoRR}, abs/1711.10228, 2017.

\bibitem{Bui2018SceneCA}
M.~Bui, S.~Albarqouni, S.~Ilic, and N.~Navab.
\newblock {Scene Coordinate and Correspondence Learning for Image-Based
  Localization}.
\newblock In {\em British Machine Vision Conference}, 2018.

\bibitem{Bujnak2010NewES}
M.~Bujnak, Z.~Kukelova, and T.~Pajdla.
\newblock {New Efficient Solution to the Absolute Pose Problem for Camera with
  Unknown Focal Length and Radial Distortion}.
\newblock In {\em Asian Conference on Computer Vision}, 2010.

\bibitem{Chen2011CityscaleLI}
D.~M. Chen, G.~Baatz, K.~K{\"o}ser, S.~S. Tsai, R.~Vedantham,
  T.~Pylv{\"a}n{\"a}inen, K.~Roimela, X.~Chen, J.~Bach, M.~Pollefeys, B.~Girod,
  and R.~Grzeszczuk.
\newblock {City-Scale Landmark Identification on Mobile Devices}.
\newblock In {\em Conference on Computer Vision and Pattern Recognition}, pages
  737--744, 2011.

\bibitem{Choudhary:2012:VPS:2403138.2403150}
S.~Choudhary and P.~J. Narayanan.
\newblock {Visibility Probability Structure from Sfm Datasets and
  Applications}.
\newblock In {\em European Conference on Computer Vision}, pages 130--143,
  2012.

\bibitem{Cummins2008FABMAPPL}
M.~J. Cummins and P.~Newman.
\newblock {FAB-MAP: Probabilistic Localization and Mapping in the Space of
  Appearance}.
\newblock {\em I. J. Robotics Res.}, 27:647--665, 2008.

\bibitem{SuperPoint}
D.~DeTone, T.~Malisiewicz, and A.~Rabinovich.
\newblock {Superpoint: Self-Supervised Interest Point Detection and
  Description}.
\newblock {\em CoRR}, abs/1712.07629, 2017.

\bibitem{Donoser2014DiscriminativeFM}
M.~Donoser and D.~Schmalstieg.
\newblock {Discriminative Feature-To-Point Matching in Image-Based
  Localization}.
\newblock In {\em Conference on Computer Vision and Pattern Recognition}, pages
  516--523, 2014.

\bibitem{Fischler1981RandomSC}
M.~A. Fischler and R.~C. Bolles.
\newblock Random sample consensus: A paradigm for model fitting with
  applications to image analysis and automated cartography.
\newblock {\em Commun. ACM}, 24:381--395, 1981.

\bibitem{RANSAC}
M.~A. Fischler and R.~C. Bolles.
\newblock {Random Sample Consensus: A Paradigm for Model Fitting with
  Applications to Image Analysis and Automated Cartography}.
\newblock {\em Commun. ACM}, 24:381--395, 1981.

\bibitem{ECBR}
H.~Germain, G.~Bourmaud, and V.~Lepetit.
\newblock {Improving Nighttime Retrieval-Based Localization}.
\newblock 2018.

\bibitem{Gordo2017EndtoEndLO}
A.~Gordo, J.~Almaz{\'a}n, J.~Revaud, and D.~Larlus.
\newblock {End-To-End Learning of Deep Visual Representations for Image
  Retrieval}.
\newblock {\em International Journal of Computer Vision}, 124:237--254, 2017.

\bibitem{Haralick1994ReviewAA}
R.~M. Haralick, C.-N. Lee, K.~Ottenberg, and M.~N{\"o}lle.
\newblock {Review and Analysis of Solutions of the Three Point Perspective Pose
  Estimation Problem}.
\newblock {\em International Journal of Computer Vision}, 13:331--356, 1994.

\bibitem{hypercolumns}
B.~Hariharan, P.~A. Arbel{\'a}ez, R.~B. Girshick, and J.~Malik.
\newblock {Hypercolumns for Object Segmentation and Fine-Grained Localization}.
\newblock In {\em Conference on Computer Vision and Pattern Recognition}, pages
  447--456, 2015.

\bibitem{ResNet}
K.~He, X.~Zhang, S.~Ren, and J.~Sun.
\newblock {Deep Residual Learning for Image Recognition}.
\newblock In {\em Conference on Computer Vision and Pattern Recognition}, pages
  770--778, 2016.

\bibitem{Heisterklaus2014ImagebasedPE}
I.~Heisterklaus, N.~Qian, and A.~Miller.
\newblock {Image-Based Pose Estimation Using a Compact 3D Model}.
\newblock In {\em International Conference on Consumer Electronics Berlin},
  pages 327--330, 2014.

\bibitem{Irschara2009FromSP}
A.~Irschara, C.~Zach, J.-M. Frahm, and H.~Bischof.
\newblock From structure-from-motion point clouds to fast location recognition.
\newblock {\em 2009 IEEE Conference on Computer Vision and Pattern
  Recognition}, pages 2599--2606, 2009.

\bibitem{Jgou2012NegativeEA}
H.~J{\'e}gou and O.~Chum.
\newblock {Negative Evidences and Co-Occurrences in Image Retrieval: the
  Benefit of PCA and Whitening}.
\newblock In {\em European Conference on Computer Vision}, 2012.

\bibitem{Jgou2011ProductQF}
H.~J{\'e}gou, M.~Douze, and C.~Schmid.
\newblock {Product Quantization for Nearest Neighbor Search}.
\newblock {\em IEEE Transactions on Pattern Analysis and Machine Intelligence},
  33:117--128, 2011.

\bibitem{Kalantidis2016CrossdimensionalWF}
Y.~Kalantidis, C.~Mellina, and S.~Osindero.
\newblock {Cross-Dimensional Weighting for Aggregated Deep Convolutional
  Features}.
\newblock In {\em European Conference on Computer Vision}, 2016.

\bibitem{Kendall2017GeometricLF}
A.~Kendall and R.~Cipolla.
\newblock {Geometric Loss Functions for Camera Pose Regression with Deep
  Learning}.
\newblock In {\em Conference on Computer Vision and Pattern Recognition}, pages
  6555--6564, 2017.

\bibitem{Kendall2015PoseNetAC}
A.~Kendall, M.~K. Grimes, and R.~Cipolla.
\newblock {Posenet: A Convolutional Network for Real-Time 6-DOF Camera
  Relocalization}.
\newblock In {\em International Conference on Computer Vision}, pages
  2938--2946, 2015.

\bibitem{P3P}
L.~Kneip, D.~Scaramuzza, and R.~Siegwart.
\newblock {A Novel Parametrization of the Perspective-Three-Point Problem for a
  Direct Computation of Absolute Camera Position and Orientation}.
\newblock In {\em Conference on Computer Vision and Pattern Recognition}, pages
  2969--2976, 2011.

\bibitem{Kobyshev2014MatchingFC}
N.~Kobyshev, H.~Riemenschneider, and L.~{Van~Gool}.
\newblock {Matching Features Correctly through Semantic Understanding}.
\newblock In {\em International Conference on 3D Vision}, pages 472--479, 2014.

\bibitem{Kukelova2013RealTimeST}
Z.~Kukelova, M.~Bujnak, and T.~Pajdla.
\newblock {Real-Time Solution to the Absolute Pose Problem with Unknown Radial
  Distortion and Focal Length}.
\newblock In {\em International Conference on Computer Vision}, pages
  2816--2823, 2013.

\bibitem{Larsson2016OutlierRF}
V.~Larsson, J.~Fredriksson, C.~Toft, and F.~Kahl.
\newblock {Outlier Rejection for Absolute Pose Estimation with Known
  Orientation}.
\newblock In {\em British Machine Vision Conference}, 2016.

\bibitem{Lepetit2008EPnPAA}
V.~Lepetit, F.~Moreno-Noguer, and P.~Fua.
\newblock Epnp: An accurate o(n) solution to the pnp problem.
\newblock {\em International Journal of Computer Vision}, 81:155--166, 2008.

\bibitem{Li:2010:LRU:1888028.1888088}
Y.~Li, N.~Snavely, and D.~P. Huttenlocher.
\newblock {Location Recognition Using Prioritized Feature Matching}.
\newblock In {\em European Conference on Computer Vision}, pages 791--804,
  2010.

\bibitem{Li2012WorldwidePE}
Y.~Li, N.~Snavely, D.~P. Huttenlocher, and P.~Fua.
\newblock {Worldwide Pose Estimation Using 3D Point Clouds}.
\newblock In {\em Large-Scale Visual Geo-Localization}, 2012.

\bibitem{Lim2012RealtimeI6}
H.~Lim, S.~N. Sinha, M.~F. Cohen, and M.~Uyttendaele.
\newblock {Real-Time Image-Based 6-DOF Localization in Large-Scale
  Environments}.
\newblock In {\em Conference on Computer Vision and Pattern Recognition}, pages
  1043--1050, 2012.

\bibitem{Liu2017EfficientG2}
L.~Liu, H.~Li, and Y.~Dai.
\newblock {Efficient Global 2D-3D Matching for Camera Localization in a
  Large-Scale 3D Map}.
\newblock In {\em International Conference on Computer Vision}, pages
  2391--2400, 2017.

\bibitem{SIFT}
D.~G. Lowe.
\newblock {Distinctive Image Features from Scale-Invariant Keypoints}.
\newblock {\em International Journal of Computer Vision}, 60(2):91--110,
  November 2004.

\bibitem{Lynen2015GetOO}
S.~Lynen, T.~Sattler, M.~Bosse, J.~A. Hesch, M.~Pollefeys, and R.~Siegwart.
\newblock {Get Out of My Lab: Large-Scale, Real-Time Visual-Inertial
  Localization}.
\newblock In {\em Robotics: Science and Systems}, 2015.

\bibitem{RobotCar}
W.~Maddern, G.~Pascoe, C.~Linegar, and P.~Newman.
\newblock {1 Year, 1000 Km: the Oxford Robotcar Dataset}.
\newblock {\em I. J. Robotics Res.}, 36:3--15, 2017.

\bibitem{McManus2014ShadyDR}
C.~McManus, W.~Churchill, W.~P. Maddern, A.~D. Stewart, and P.~Newman.
\newblock {Shady Dealings: Robust, Long-Term Visual Localisation Using
  Illumination Invariance}.
\newblock In {\em IEEE International Conference on Robotics and Automation},
  pages 901--906, 2014.

\bibitem{Middelberg2014Scalable6L}
S.~Middelberg, T.~Sattler, O.~Untzelmann, and L.~Kobbelt.
\newblock {Scalable 6-DOF Localization on Mobile Devices}.
\newblock In {\em European Conference on Computer Vision}, 2014.

\bibitem{Noh2017LargeScaleIR}
H.~Noh, A.~Araujo, J.~Sim, T.~Weyand, and B.~Han.
\newblock Large-scale image retrieval with attentive deep local features.
\newblock {\em 2017 IEEE International Conference on Computer Vision (ICCV)},
  pages 3476--3485, 2017.

\bibitem{Ono2018LFNetLL}
Y.~Ono, E.~Trulls, P.~Fua, and K.~M. Yi.
\newblock {{LF-Net}: Learning Local Features from Images}.
\newblock In {\em NeurIPS}, 2018.

\bibitem{Radenovic2018RevisitingOA}
F.~Radenovic, A.~Iscen, G.~Tolias, Y.~S. Avrithis, and O.~Chum.
\newblock {Revisiting Oxford and Paris: Large-Scale Image Retrieval
  Benchmarking}.
\newblock {\em CoRR}, abs/1803.11285, 2018.

\bibitem{GeM}
F.~Radenovic, G.~Tolias, and O.~Chum.
\newblock {Fine-Tuning CNN Image Retrieval with No Human Annotation}.
\newblock {\em IEEE Transactions on Pattern Analysis and Machine Intelligence},
  2018.

\bibitem{Razavian2014VisualIR}
A.~S. Razavian, J.~Sullivan, A.~Maki, and S.~Carlsson.
\newblock {Visual Instance Retrieval with Deep Convolutional Networks}.
\newblock {\em CoRR}, abs/1412.6574, 2014.

\bibitem{HFNet}
P.-E. Sarlin, C.~Cadena, R.~Siegwart, and M.~Dymczyk.
\newblock {From Coarse to Fine: Robust Hierarchical Localization at Large
  Scale}.
\newblock 2019.

\bibitem{Sarlin2018LeveragingDV}
P.-E. Sarlin, F.~Debraine, M.~Dymczyk, and R.~Siegwart.
\newblock Leveraging deep visual descriptors for hierarchical efficient
  localization.
\newblock In {\em CoRL}, 2018.

\bibitem{Sattler2015HyperpointsAF}
T.~Sattler, M.~Havlena, F.~Radenovic, K.~Schindler, and M.~Pollefeys.
\newblock {Hyperpoints and Fine Vocabularies for Large-Scale Location
  Recognition}.
\newblock In {\em International Conference on Computer Vision}, pages
  2102--2110, 2015.

\bibitem{Sattler2016LargeScaleLR}
T.~Sattler, M.~Havlena, K.~Schindler, and M.~Pollefeys.
\newblock {Large-Scale Location Recognition and the Geometric Burstiness
  Problem}.
\newblock In {\em Conference on Computer Vision and Pattern Recognition}, pages
  1582--1590, 2016.

\bibitem{ActiveSearch}
T.~Sattler, B.~Leibe, and L.~Kobbelt.
\newblock {Improving Image-Based Localization by Active Correspondence Search}.
\newblock In {\em European Conference on Computer Vision}, 2012.

\bibitem{Sattler2017EfficientE}
T.~Sattler, B.~Leibe, and L.~Kobbelt.
\newblock {Efficient \& Effective Prioritized Matching for Large-Scale
  Image-Based Localization}.
\newblock {\em IEEE Transactions on Pattern Analysis and Machine Intelligence},
  39:1744--1756, 2017.

\bibitem{6DOFBenchmark}
T.~Sattler, W.~Maddern, C.~Toft, A.~Torii, L.~Hammarstrand, E.~Stenborg,
  D.~Safari, M.~Okutomi, M.~Pollefeys, J.~Sivic, F.~Kahl, and T.~Pajdla.
\newblock {Benchmarking 6DOF Outdoor Visual Localization in Changing
  Conditions}.
\newblock In {\em Conference on Computer Vision and Pattern Recognition}, June
  2018.

\bibitem{Sattler2014OnSF}
T.~Sattler, C.~Sweeney, and M.~Pollefeys.
\newblock {On Sampling Focal Length Values to Solve the Absolute Pose Problem}.
\newblock In {\em European Conference on Computer Vision}, 2014.

\bibitem{sattler:hal-01513083}
T.~Sattler, A.~Torii, J.~Sivic, M.~Pollefeys, H.~Taira, M.~Okutomi, and
  T.~Pajdla.
\newblock {Are Large-Scale 3D Models Really Necessary for Accurate Visual
  Localization?}
\newblock In {\em Conference on Computer Vision and Pattern Recognition},
  page~10, July 2017.

\bibitem{COLMAP1}
J.~L. Sch{\"o}nberger and J.-M. Frahm.
\newblock {Structure-From-Motion Revisited}.
\newblock In {\em Conference on Computer Vision and Pattern Recognition}, pages
  4104--4113, 2016.

\bibitem{Schnberger2017SemanticVL}
J.~L. Sch{\"o}nberger, M.~Pollefeys, A.~Geiger, and T.~Sattler.
\newblock {Semantic Visual Localization}.
\newblock {\em CoRR}, abs/1712.05773, 2017.

\bibitem{COLMAP2}
J.~L. Sch{\"o}nberger, E.~Zheng, J.-M. Frahm, and M.~Pollefeys.
\newblock {Pixelwise View Selection for Unstructured Multi-View Stereo}.
\newblock In {\em European Conference on Computer Vision}, 2016.

\bibitem{VGG16}
K.~Simonyan and A.~Zisserman.
\newblock {Very Deep Convolutional Networks for Large-Scale Image Recognition}.
\newblock {\em CoRR}, abs/1409.1556, 2015.

\bibitem{pub.1046137732}
G.~Singh and J.~Košecká.
\newblock {\em {Semantically Guided Geo-Location and Modeling in Urban
  Environments}}, pages 101--120.
\newblock 2016.

\bibitem{CSL}
L.~Sv\"{a}rm, O.~Enqvist, F.~Kahl, and M.~Oskarsson.
\newblock {City-Scale Localization for Cameras with Known Vertical Direction}.
\newblock {\em IEEE Transactions on Pattern Analysis and Machine Intelligence},
  39(7):1455--1461, 7 2017.

\bibitem{Svrm2014AccurateLA}
L.~Sv\"{a}rm, O.~Enqvist, M.~Oskarsson, and F.~Kahl.
\newblock {Accurate Localization and Pose Estimation for Large 3D Models}.
\newblock In {\em Conference on Computer Vision and Pattern Recognition}, pages
  532--539, 2014.

\bibitem{Taira2018InLocIV}
H.~Taira, M.~Okutomi, T.~Sattler, M.~Cimpoi, M.~Pollefeys, J.~Sivic, T.~Pajdla,
  and A.~Torii.
\newblock {Inloc: Indoor Visual Localization with Dense Matching and View
  Synthesis}.
\newblock {\em CoRR}, abs/1803.10368, 2018.

\bibitem{SMC}
C.~Toft, E.~Stenborg, L.~Hammarstrand, L.~Brynte, M.~Pollefeys, T.~Sattler, and
  F.~Kahl.
\newblock {Semantic Match Consistency for Long-Term Visual Localization}.
\newblock In {\em European Conference on Computer Vision}, 09 2018.

\bibitem{Tolias2015ParticularOR}
G.~Tolias, R.~Sicre, and H.~J{\'e}gou.
\newblock {Particular Object Retrieval with Integral Max-Pooling of CNN
  Activations}.
\newblock {\em CoRR}, abs/1511.05879, 2015.

\bibitem{DenseVLAD}
A.~Torii, R.~Arandjelovic, J.~Sivic, M.~Okutomi, and T.~Pajdla.
\newblock {24/7 Place Recognition by View Synthesis}.
\newblock {\em IEEE Transactions on Pattern Analysis and Machine Intelligence},
  40:257--271, 2015.

\bibitem{Walch2017ImageBasedLU}
F.~Walch, C.~Hazirbas, L.~Leal-Taix{\'e}, T.~Sattler, S.~Hilsenbeck, and
  D.~Cremers.
\newblock {Image-Based Localization Using LSTMs for Structured Feature
  Correlation}.
\newblock In {\em International Conference on Computer Vision}, pages 627--637,
  2017.

\bibitem{Yi16}
K.~M. Yi, E.~Trulls, V.~Lepetit, and P.~Fua.
\newblock {LIFT: Learned Invariant Feature Transform}.
\newblock In {\em European Conference on Computer Vision}, 2016.

\bibitem{Zamir2010AccurateIL}
A.~R. Zamir and M.~Shah.
\newblock {Accurate Image Localization Based on Google Maps Street View}.
\newblock In {\em European Conference on Computer Vision}, 2010.

\bibitem{Zeisl2015CameraPV}
B.~Zeisl, T.~Sattler, and M.~Pollefeys.
\newblock {Camera Pose Voting for Large-Scale Image-Based Localization}.
\newblock In {\em International Conference on Computer Vision}, pages
  2704--2712, 2015.

\bibitem{Zhang2006ImageBL}
W.~Zhang and J.~Kosecka.
\newblock {Image Based Localization in Urban Environments}.
\newblock {\em International Symposium on 3D Data Processing, Visualization,
  and Transmission}, pages 33--40, 2006.

\end{thebibliography}
}

\end{document}